\documentclass[letterpaper, 10 pt, conference]{ieeeconf}  

\IEEEoverridecommandlockouts                              

\usepackage{graphicx} 
\usepackage{color} 
\usepackage{amsmath} 
\usepackage{amssymb}  
\usepackage{pifont}
\usepackage{subcaption}
\usepackage[colorlinks=true, linkcolor=blue, citecolor=green, urlcolor=red]{hyperref}

\title{\LARGE \bf
BODex: Scalable and Efficient Robotic Dexterous Grasp Synthesis \\ Using Bilevel Optimization
}

\author{Jiayi Chen$^{1, 2*}$, Yubin Ke$^{1, 2*}$ and He Wang$^{1, 2, 3\dagger}$
\thanks{$^{1}$Peking University. $^{2}$Galbot. $^{3}$Beijing Academy of Artificial Intelligence.}%
\thanks{$^*$Equal contribution. $^\dagger$Corresponding author: \href{mailto:hewang@pku.edu.cn}{hewang@pku.edu.cn}.}
}

\begin{document}

\newcommand{\jiayi}[1]{{\color{red}[Jiayi: #1]}}
\newcommand{\yb}[1]{{\color{blue}[Yubing: #1]}}

\maketitle
\thispagestyle{empty}
\pagestyle{empty}

\begin{abstract}

Robotic dexterous grasping is important for interacting with the environment. To unleash the potential of data-driven models for dexterous grasping, a large-scale, high-quality dataset is essential. While gradient-based optimization offers a promising way for constructing such datasets, previous works suffer from limitations, such as inefficiency, strong assumptions in the grasp quality energy, or limited object sets for experiments. 
Moreover, the lack of a standard benchmark for comparing different methods and datasets hinders progress in this field.
To address these challenges, we develop a highly efficient synthesis system and a comprehensive benchmark with MuJoCo for dexterous grasping. We formulate grasp synthesis as a bilevel optimization problem, combining a novel lower-level quadratic programming (QP) with an upper-level gradient descent process. By leveraging recent advances in CUDA-accelerated robotic libraries and GPU-based QP solvers, our system can parallelize thousands of grasps and synthesize over $49$ grasps per second on a single 3090 GPU. Our synthesized grasps for Shadow, Allegro, and Leap hands all achieve a success rate above $75\%$ in simulation, with a penetration depth under $1$ mm, outperforming existing baselines on nearly all metrics. Compared to the previous large-scale dataset, DexGraspNet, our dataset significantly improves the performance of learning models, with a success rate from around $40\%$ to $80\%$ in simulation. Real-world testing of the trained model on the Shadow Hand achieves an $81\%$ success rate across 20 diverse objects. The codes and datasets are released on our project page: \href{https://pku-epic.github.io/BODex}{https://pku-epic.github.io/BODex}.

\end{abstract}

\section{Introduction}

Robotic dexterous grasping is foundational to interacting with the environment and thus an important research topic. While large-scale data collection and learning-based methods have achieved significant success in parallel gripper grasping~\cite{fang2020graspnet, fang2023anygrasp}, their potential for dexterous hands remains largely unexplored, partly due to the increased difficulty of data collection. Unlike parallel grippers, dexterous hands often have over 20 degrees of freedom (DoF), thus greatly reducing the effectiveness of directly sampling grasp poses.

Although gradient-based optimization has been explored recently as a promising approach to scaling up the grasp data for dexterous hands, previous methods face several limitations. Some~\cite{liu2021synthesizing, wang2023dexgraspnet} rely on strong assumptions in the grasp quality energy, such as equal contact forces and no friction, while others~\cite{li2023frogger, chen2024springgrasp} only study on a limited set of objects. Moreover, the synthesis speed of previous works is quite slow. In addition, differences in robot hands, simulators, and evaluation metrics across studies make comparison difficult.

To address these challenges, we develop an efficient grasp synthesis system and a comprehensive benchmark. We formulate grasp synthesis as a bilevel optimization problem: the lower-level quadratic programming (QP) determines the optimal force combination for each contact at the current hand pose to achieve a desired wrench, without any assumption, while the upper-level process performs gradient descent on the hand pose to minimize the difference between the desired wrench and the best-applied wrench, as determined by the lower-level QP.
Our system can also synthesize pre-grasp poses that maintain a certain distance from the object, aiding in planning collision-free hand-arm trajectories and controlling the hand to apply force on the object.

\begin{figure}
    \centering
    \includegraphics[width=0.8\columnwidth]{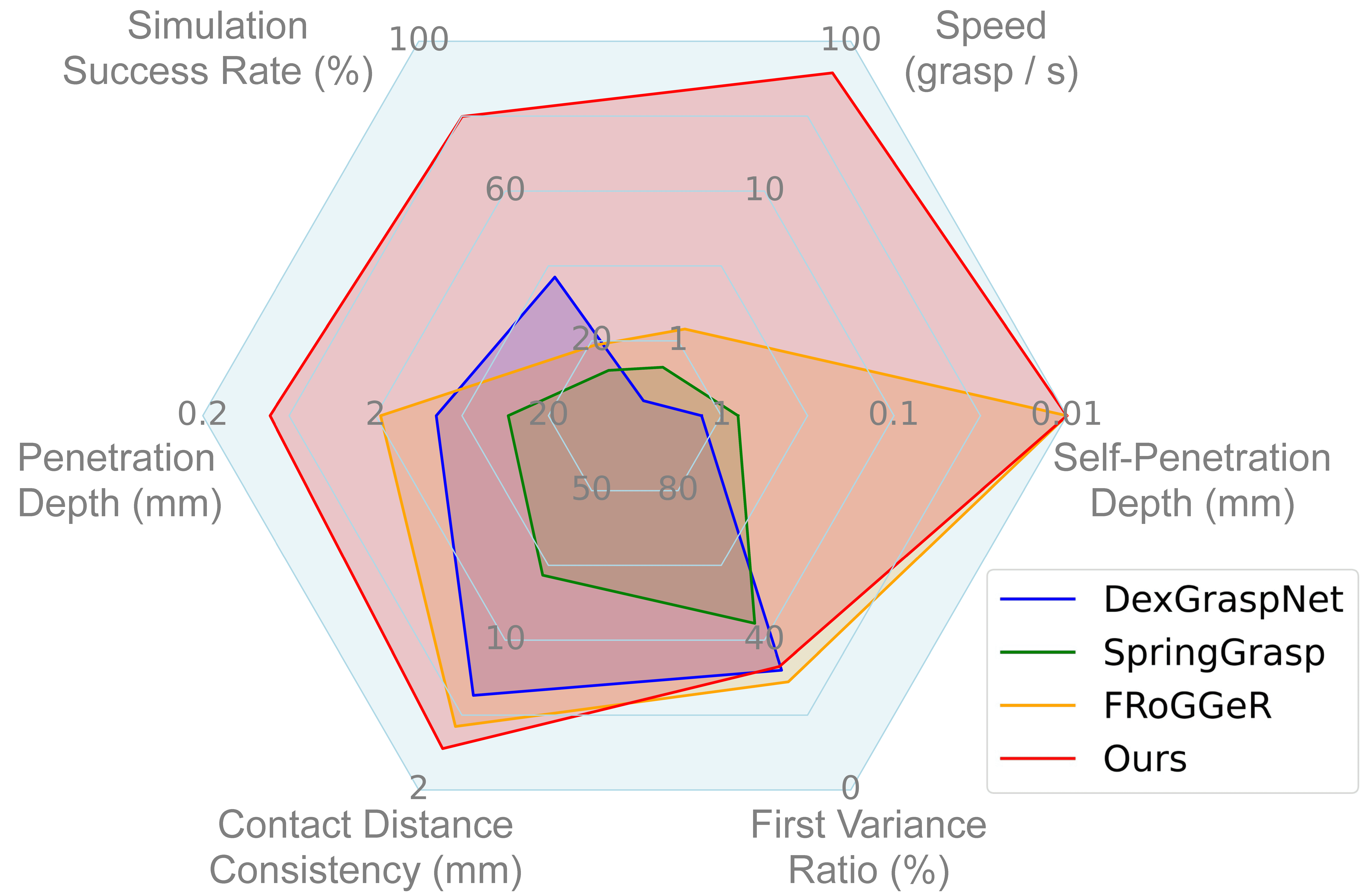}
    \caption{\textbf{Comparison with analytic-based dexterous grasp synthesis baselines on Allegro Hand.} Our pipeline significantly outperforms baselines on almost all metrics, especially on the most important two, simulation success rate and speed. }
    \label{fig:teaser}
    \vspace{-4mm}
\end{figure}

To accelerate the system and enable large-scale parallelization, we leverage recent advances in the CUDA-accelerated robotics library, cuRobo~\cite{curobo_report23}, and the GPU-based QP solver, ReLU-QP~\cite{bishop2024relu}. To integrate these tools into our system, we first propose a coarse-to-fine strategy to address imprecise contact issues caused by sphere approximation in cuRobo. We also implement a batched version of ReLU-QP to solve multiple QPs in parallel on a GPU, achieving a 10x speedup compared to CPU solvers like ProxQP~\cite{bambade2023proxqp} and OSQP~\cite{stellato2020osqp}.

\begin{table*}[]
\vspace{2mm}
    \centering
    \begin{tabular}{c|c|c|c|c|c|c|c|c}
        Dataset & Hand & Sim./Real & Objects & Grasps & Table & Pre-grasps & Collision-free trajectory & Method  \\
        \hline
        \hline 
        DDGdata~\cite{liu2020deep} & Shadow & Sim. & 565
 & 6.9k & \ding{55} & \ding{55} & \ding{55} & \textit{GraspIt!} \\
 
       DexGraspNet~\cite{wang2023dexgraspnet}  & Shadow & Sim. & 5355 & 1.32M & \ding{55} & \ding{55} & \ding{55} & Optimization \\
       
       GenDexGrasp~\cite{li2023gendexgrasp} & Multiple & Sim. & 58 & 436k & \ding{55} & \ding{55} & \ding{55} &  Optimization  \\
       
       RealDex~\cite{liu2024realdex} & Shadow & Real & 52 & 59k & \checkmark & \ding{55} & 2630 & Teleoperation  \\
       \hline
       BODex-Floating & Shadow, Allegro, Leap & Sim. & 2397 & 4.09M & \ding{55} & \checkmark & \ding{55} & Optimization \\
      BODex-Tabletop & UR10e+Shadow & Sim. & 2397 & 3.08M & \checkmark & \checkmark & 3.08M & Optimization \\
    \end{tabular}
    \caption{\textbf{Robotic dexterous grasping dataset statistic.} All of our grasps and trajectories have been validated in MuJoCo.}
    \label{tab:my_label}
    \vspace{-4mm}
\end{table*}

Finally, we establish a benchmark with MuJoCo~\cite{todorov2012mujoco} to compare various analytic-based grasp synthesis pipelines, grasp energy functions, and learning-based methods. As shown in Fig.~\ref{fig:teaser}, our system significantly outperforms baselines across nearly all metrics, achieving a 50x speedup in synthesizing higher-quality grasps. Compared to previous grasp energies, our QP-based energy does not rely on assumptions about contact forces and friction, resulting in a higher success rate and better correlation with simulation outcomes (i.e. a grasp with lower energy is more likely to succeed in simulation). The learning-based method trained on our dataset also greatly outperforms the same model trained on the previous large-scale dataset, DexGraspNet, improving success rates from around $40\%$ to $80\%$. Real-world testing of the trained model on the Shadow Hand achieves an $81\%$ success rate across $20$ diverse objects.

In summary, our contributions are: (1) a GPU-based efficient grasp synthesis system using bilevel optimization; (2) a large-scale, high-quality dexterous grasp dataset that enables a better learning model; and (3) a reproducible benchmark for grasp synthesis with the MuJoCo simulator.

\section{Related Work}

\textbf{Analytic-based dexterous grasp synthesis} methods are often used for constructing datasets to train learning-based models, as analytic-based methods typically rely on complete object geometry, which is difficult to obtain in real-time but available in offline 3D assets. These methods commonly focus on synthesizing force-closure grasps that can resist any external wrench applied to the object. The most popular metric for evaluating force closure is the Q1 metric~\cite{ferrari1992planning}, which measures the radius of the largest inscribed ball within the Grasp Wrench Space (GWS) — the set of all wrenches that the hand can apply to the object. Early approaches, e.g. \textit{GraspIt!}~\cite{miller2004graspit}, use sampling-based methods to find grasps with a high Q1 metric but are inefficient for high-DoF hands.

More recent work has explored gradient-based optimization for grasp synthesis. DFC~\cite{liu2021synthesizing, li2023gendexgrasp} introduces a differentiable force closure energy that aims to include the origin within the GWS, under the assumptions of no friction and equal contact forces. DexGraspNet~\cite{wang2023dexgraspnet} accelerates the pipeline of DFC and generates a large-scale grasp dataset for over 5000 objects, but its speed and data quality still need improvement. Grasp'D~\cite{turpin2022grasp} and Fast-Grasp'D~\cite{turpin2023fast} explore the use of differentiable simulators for grasp synthesis. FRoGGeR~\cite{li2023frogger} and SpringGrasp~\cite{chen2024springgrasp} propose novel energies for optimization but study only a limited set of objects. TaskDexGrasp~\cite{chen2023task} extends the formulation to both force-closure and non-force-closure grasps. 

The grasp energy proposed in \cite{wu2022learning} is the most similar one to ours, utilizing QP and relaxing the assumptions of DFC. However, ~\cite{wu2022learning} only uses it for post-processing network outputs and not for large-scale dataset synthesis. We provide comparisons of our synthesis pipeline and energy function against previous works to show our effectiveness.

\textbf{Learning-based dexterous grasp synthesis} methods support inference from partial visual input, making them more suitable as policies for real-time execution. Supervised learning methods~\cite{jiang2021graspTTA, chen2022learning, xu2023unidexgrasp} rely on offline datasets for training and often utilize generative models such as CVAE~\cite{kingma2013auto}, diffusion models~\cite{ho2020denoising}, and normalizing flows~\cite{rezende2015variational}. Some approaches~\cite{wan2023unidexgrasp++} have also explored reinforcement learning for dexterous grasping, though Sim2Real transfer of these policies remains an open challenge. We benchmark several supervised learning methods in simulation, showing that models trained on our dataset outperform those trained on the previous large-scale dataset, DexGraspNet.

\textbf{Bilevel optimization for grasp synthesis}~\cite{li2023frogger, wu2022learning} involves an upper-level optimization process that solves a lower-level optimization problem at each iteration. This formulation nests two optimization problems and differs from sequential optimization~\cite{schulman2014motion}. It typically incurs higher computational costs than methods whose grasp energies do not require optimization. To address this, we implement a GPU-based QP solver~\cite{bishop2024relu} capable of solving lower-level QPs in parallel, preventing it from becoming a speed bottleneck.

\section{Preliminaries}

This section introduces the most popular contact model, point contact with friction. Consider an object $O$ is grasped by a robot hand with $m$ contacts. For each contact $i \in \{1, \cdots, m\}$, let $\mathbf{p}_i\in \mathbb{R}^3$ be the contact position, $\mathbf{n}_i\in \mathbb{R}^3$ the inward-pointing surface unit normal, and $\mathbf{d}_i\in \mathbb{R}^3$ and $\mathbf{e}_i\in \mathbb{R}^3$ two unit tangent vectors satisfying $\mathbf{n}_i = \mathbf{d}_i \times \mathbf{e}_i$, all of which are defined in the object coordinate frame with the gravity center as the origin. The contact model is:
\begin{align}
\label{eq: F_pcf}
\mathcal{F}_i & = \left\{\mathbf{f}_i\in\mathbb{R}^3~|~0 \leq f_{i,1} \leq 1, f_{i,2}^2+f_{i,3}^2 \leq \mu^2 f_{i,1}^2 \right\} 
\\
\label{eq: G_pcf}
\mathbf{G}_i & = 
\begin{bmatrix}
    \mathbf{n}_i & \mathbf{d}_i & \mathbf{e}_i \\
    \mathbf{p}_i \times \mathbf{n}_i & 
    \mathbf{p}_i \times \mathbf{d}_i & 
    \mathbf{p}_i \times \mathbf{e}_i \\
\end{bmatrix} \in \mathbb{R}^{6\times3}
\end{align}
where $\mu$ is the friction coefficient, $\mathcal{F}_i$ contains all possible forces that can be generated by contact $i$, and the matrix $\mathbf{G}_i$ maps the contact force $\mathbf{f}_i$ to a wrench $\mathbf{w}_i = \mathbf{G}_i \mathbf{f}_i$.

\section{Method}

\subsection{Bilevel Optimization Formulation for Grasp Synthesis}
\label{sec:qp}

We formulate the dexterous grasp synthesis as a nonlinear bilevel optimization program as follows:
\begin{align}
    \underset{\mathbf{x}, \mathbf{y}_j, j\in\{1,...,s\}}{\text{minimize}}~~ & \sum_{j=1}^{s} Q_j(\mathbf{x})  \label{eq:upper-level} \\
    \text{s.t.}~~& \mathbf{x}_{min}\leq\mathbf{x}\leq\mathbf{x}_{max} \label{constraint_limit} \\
    & \mathbf{c}_{i,w}=\text{FK}(\mathbf{x}, \mathbf{c}_{i,l}) \in \delta O,i \in \{1,...,m\} \label{constraint_contact} \\
    & \text{No (hand-hand/hand-object) collision.} \label{constraint_collision}
\end{align}
The object mesh $O$ and the expected hand contact points $\{\mathbf{c}_{i,l}\}$ in the link frame are the input, while the output is the grasp pose $\mathbf{x}=[\mathbf{r}, \mathbf{t}, \mathbf{q}]\in\mathbb{R}^{9+3+n}$, including root rotation, translation, and $n$ joint angles. 
Constraint~\ref{constraint_limit} ensures the pose within specified ranges. Constraint~\ref{constraint_contact} requires the hand points $\mathbf{c}_{i,w}$ in the world frame to contact the object surface, where FK is forward kinematics. Eq.~\ref{eq:upper-level} is the grasp energy for the upper-level problem, where each $Q_j$ is a lower-level QP as:
\begin{align}
   Q_j(\mathbf{x}) \triangleq ~~&\underset{\mathbf{y}_j}{\text{min}}~~\|\beta \mathbf{t}_j - \Sigma_{i=1}^m\mathbf{G}_i\mathbf{f}_{j,i}\|^2 \label{eq: qp} \\
    &\text{s.t.}~~~~\mathbf{f}_{j,i} \in \mathcal{F}_i, ~~i\in\{1,...,m\} \label{constraint_friction} \\
    & ~~~~~~~\Sigma_{i=1}^m f_{j,i,1} \ge \gamma  \label{contraint_f_sum}  
\end{align}
where $\mathbf{y}_j=[\mathbf{f}_{j,1},...,\mathbf{f}_{j,m}]\in\mathbb{R}^{3m}$, $\mathbf{t}_j$ is a given unit vector indicating the desired wrench direction, $\beta$ and $\gamma$ are two positive hyperparameters. 

By finding the optimal contact forces $\mathbf{y}_j$ for a desired wrench $\beta\mathbf{t}_j$, the grasp energy $Q_j(\mathbf{x})$ measures the difference between the desired wrench and the best-applied wrench $\Sigma_{i=1}^m\mathbf{G}_i\mathbf{f}_{j,i}$ at the grasp pose $\mathbf{x}$. To get a better grasp pose, the upper-level process performs gradient descent on $Q_j$, which is differentiable to $\mathbf{x}$ because $\mathbf{G}_i$ is differentiable to $\mathbf{x}$. The final grasp energy sums up several $Q_j$ with different $\mathbf{t}_j$ to encourage the applicable wrenches in multiple directions. For a force-closure grasp, $\{\mathbf{t}_j\}$ is set to be the six unit vectors along the positive and negative 3D force axes, with zero 3D torques, e.g., $[1, 0, 0, 0, 0, 0]$ and $[-1, 0, 0, 0, 0, 0]$. If a grasp can resist these six $\mathbf{t}_j$, it can resist the object gravity in any direction by a linear combination of these $\mathbf{t}_j$. Our formulation is also more flexible in customizing desired wrenches than the predefined primitive in TDG~\cite{chen2023task}.

Contraint~\ref{contraint_f_sum} is used to avoid the trivial solution of $\mathbf{y}_{j}=\mathbf{0}$, which makes $Q_j$ non-differentiable to $\mathbf{G}_i$. Constraint~\ref{constraint_friction} ensures the contact forces within the friction cones defined in Eq.~\ref{eq: F_pcf}. 
To reduce the complexity, we approximate the elliptic friction cones with 8-vertex pyramidal cones, transforming the original quadratic constrained quadratic program (QCQP) into a linear constrained quadratic program (LCQP).

\subsection{Solving the Bilevel Optimization for Grasp Synthesis}
\label{sec:pipeline}

\begin{figure}[t]
    \centering
    \vspace{2mm}
    \includegraphics[width=0.98\columnwidth]{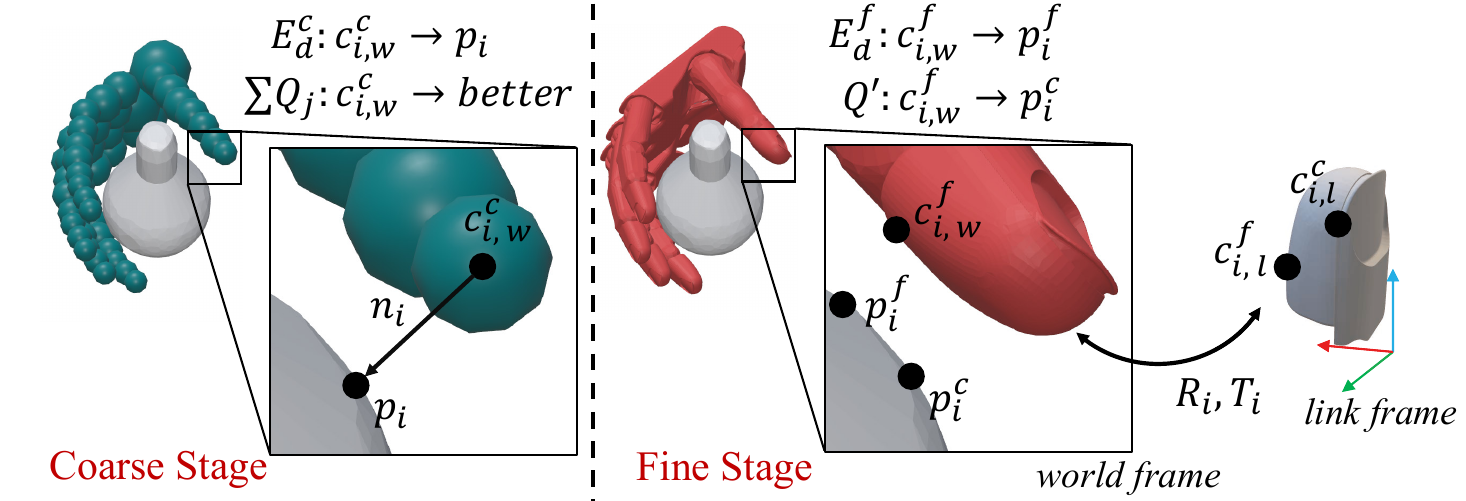}
    \caption{\textbf{Coarse-to-fine Strategy.}}
    \label{fig:coarse_to_fine}
    \vspace{-4mm}
\end{figure}

In each iteration of the upper-level optimization, the grasp pose $\mathbf{x}$ is used to compute the transformation of each hand link, represented as $\mathbf{R}_i$ and $\mathbf{T}_i$, by forward kinematics (FK). This yields the expected hand contact points in the world frame, given by $\mathbf{c}_{i, w}=\mathbf{R}_i\mathbf{c}_{i,l} + \mathbf{T}_i$. Next, the object points $\mathbf{p}_i$ and normals $\mathbf{n}_i$ are calculated by nearest-point query, as shown in the left of Fig.~\ref{fig:coarse_to_fine}. They are then used to construct the grasp matrix $\mathbf{G}_i$ in Eq.~\ref{eq: G_pcf} and the QPs in Eq.~\ref{eq: qp}. 

To efficiently solve the lower-level QPs, we implement a batched version of \texttt{ReLU-QP}~\cite{bishop2024relu}, a PyTorch-based ADMM solver that enables the parallel solving of multiple QPs with the same format on a GPU. For common constraints in the upper-level problem, we utilize the corresponding energy functions in cuRobo~\cite{curobo_report23}, such as the joint limitation energy for Constraint~\ref{constraint_limit}, and self-penetration and inter-penetration energies for Constraint~\ref{constraint_collision}. To integrate cuRobo into our system, we made necessary modifications, such as adding support for a floating base in both the FK and the optimizer. The floating base represents the 6-DoF state of the robot's root, which is one of the optimizable variables in our system.

\subsection{Coarse-to-fine Contact Modeling} 
\label{sec:coarse-to-fine}

To balance speed and accuracy, we propose a coarse-to-fine strategy for modeling contact in Constraint~\ref{constraint_contact}. As shown in Fig.~\ref{fig:coarse_to_fine}, the coarse stage approximates the robot's geometry with spheres, similar to cuRobo, allowing for fast nearest-point queries. However, this approximation lacks sufficient accuracy for grasp synthesis, particularly with small or thin objects. The fine stage uses collision meshes for precise contact modeling, but it is computationally slower.

In the first coarse stage, we set $\mathbf{c}_{i,l}$ as the center of the first sphere at each fingertip and define a distance energy as $E^c_d = \sum_{i=1}^m(\|\mathbf{c}_{i,w} - \mathbf{p}_i \| - \alpha)^2$, where $\alpha$ is the radius of the sphere. At this stage, the derivative of the nearest-point query is approximated using finite differences. 

In the second fine stage, we use the GJK algorithm~\cite{gilbert1988fast} to find the nearest points $\mathbf{c}_{i,w}^f$ on each fingertip and $\mathbf{p}_i^f$ on the object. Due to the non-differentiability of the GJK algorithm and the heavy computational cost of finite differences, $\mathbf{c}_{i,w}^f$ and $\mathbf{p}_i^f$ are not differentiable with respect to $\mathbf{x}$, making $Q_j$ non-differentiable as well. To address this, we define the distance energy $E_d^f$ and an alternative grasp energy $Q'$ as:
\begin{equation}
    E_d^f = \Sigma_{i=1}^m\|\mathbf{c}_{i,w}^{f'} - \mathbf{p}_i^f \|^2,~~Q' = \Sigma_{i=1}^m\|\mathbf{c}_{i,w}^{f'} - \mathbf{p}_i^{c} \|^2 
\end{equation}
\begin{equation}
\label{eq:trick}
\mathbf{c}_{i,w}^{f'}=\mathbf{R}_i\texttt{Detach}(\mathbf{c}_{i,l}^f) + \mathbf{T}_i,~\mathbf{c}_{i,l}^f=\mathbf{R}_i^{-1}(\mathbf{c}_{i,w}^f-\mathbf{T}_i) 
\end{equation}
where $\mathbf{p}_i^{c}$ is the object points obtained at the end of the coarse stage. By detaching the gradient of $\mathbf{c}_{i,l}^f$, we make $\mathbf{c}_{i,w}^{f'}$ differentiable to $\mathbf{R}_i$ and $\mathbf{T}_i$, enabling upper-level optimization. 

To accelerate computation, we use \texttt{Coal}~\cite{Pan_Coal_-_An_2025}, a state-of-the-art library implementing the GJK algorithm, along with \texttt{OpenMP} for parallelization. Since the GJK algorithm is limited to convex meshes, the object is decomposed into multiple convex parts. To reduce unnecessary computations, we introduce a broad-phase step that calculates the distance between the oriented bounding box (OBB) of each object part and the bounding sphere of the fingertip collision mesh. Only object parts with OBB-to-sphere distances less than the distance from the fingertip's sphere approximation to the entire object are considered for the GJK algorithm.

\begin{figure}
    \centering
    \vspace{2mm}
    \includegraphics[width=0.9\columnwidth]{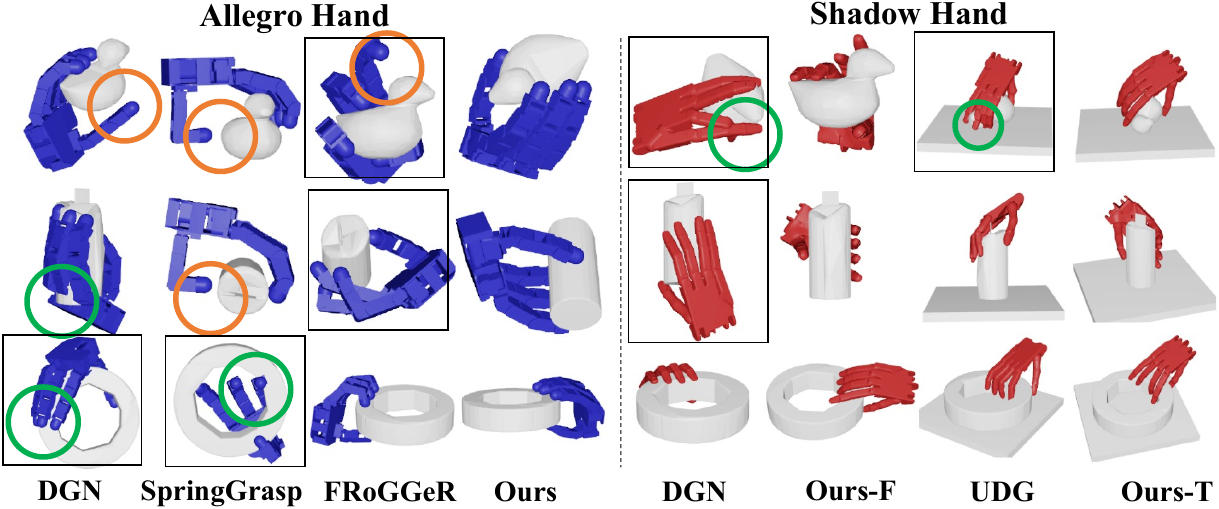}
    \caption{\textbf{Visualization of Randomly Selected Grasps.} Previous analytic-based synthesis methods show more penetration (green circles), with fingers often not contact the object (orange circles) and some unnatural poses (black boxes). 
    }
    \label{fig:sim gallery}
    \vspace{-4mm}
\end{figure}

\subsection{Collision-Free Hand-Arm Trajectory Synthesis}
\label{sec:pregrasp and trajectory}

Our system can also synthesize collision-free trajectories for each grasp pose $\mathbf{x}$ using cuRobo’s interface. However, directly planning with $\mathbf{x}$ as the target is infeasible because it involves contact with the object. Moreover, controlling the hand to reach $\mathbf{x}$ does not allow to apply force on the object.

To address these challenges, we synthesize a pre-grasp pose $\mathbf{x}_p$ that maintains a minimum distance of 1 cm from the object, achieved by reducing 1 cm when calculating the distance between the hand and object points. This pre-grasp pose also helps control the hand to apply force. Specifically, we define a squeeze pose $\mathbf{x}_s = 2\mathbf{x} - \mathbf{x}_p$ as the target for execution, both in simulation and in the real world.

In summary, our optimization consists of three stages: 
\begin{itemize}
    \item \textbf{Coarse stage} optimizes 300 iterations using collision spheres, with the nearest distance reduced by $1$ cm.
    \item \textbf{Fine stage} replaces collision spheres with meshes and adjusts the energy $Q_j$ to $Q'$, as detailed in Sec.~\ref{sec:coarse-to-fine}, for another 100 iterations to get the pre-grasp pose $\mathbf{x}_p$.
    \item \textbf{Final stage} gets the grasp pose $\mathbf{x}$ by 100 more steps, similar to \textbf{Fine stage} but without the distance reduction.
\end{itemize}

\section{Experiments}

\subsection{Simulation Environment and Object Assets}

We use the widely-adopted MuJoCo~\cite{todorov2012mujoco} simulator due to its high performance in handling contacts and constraints, as well as its stability and reproducibility. Two popular dexterous hands,  Shadow Hand and Allegro Hand, are used, whose assets are sourced from MuJoCo Menagerie~\cite{menagerie2022github}. To ensure comparability with previous datasets and baselines, we evaluate using a floating hand without a robot arm.

Most parameters of the simulator and robot assets are retained as default, with only minimal adjustments: we set \textit{noslip\_iterations} to $2$ to prevent slow slippage, a characteristic of MuJoCo for solving inverse dynamics that is irrelevant to our task. The friction model uses a tangential coefficient of $0.6$ and a torsional coefficient of $0.02$, as recommended in~\cite{xydas1999modeling}. Gravity uses $9.8\text{m/s}^2$. The object mass is set to $30\text{g}$.

Object assets are taken from DexGraspNet~\cite{wang2023dexgraspnet}. All objects are pre-processed and normalized so that the diagonal of their bounding boxes measures $2m$. To address the issue of objects being too flat to grasp on a table, we filtered out those with a shortest bounding box edge less than $0.2m$, resulting in $2,397$ valid objects. Each object is then rescaled to four different sizes: $0.06$, $0.08$, $0.10$, and $0.12$, resulting in a total of $9,588$ scaled objects for grasping.

\subsection{Evaluation Metrics}

The following metrics are used to evaluate the synthesis pipeline and grasp quality. These metrics should be considered together, since either metric alone may be misleading.

\textbf{Simulation Success Rate (SSR)} (unit: $\%$) represents the percentage of successful grasps in simulation. To perform a grasp, the hand is initialized to the pre-grasp pose $\mathbf{x}_p$ and moves to the squeezed pose $\mathbf{x}_s$. Then, the object's gravity is applied and we check whether the deviation in the object's translation and rotation angle remains within $5\text{cm}$ and $15^\circ$, respectively, for more than $3$ seconds. Each grasp is tested across six orthogonal gravity directions and is regarded as a success only if it succeeds in all directions. Our environment evaluates $30.6$ grasps per second with $60$ threads on CPUs. 

\textbf{Speed (S)} (unit: $grasp/s$) measures the number of grasps synthesized per second. Our speed is tested on an NVIDIA GeForce RTX 3090 GPU with Intel Xeon Platinum 8255C CPUs, while the speed of baselines came from their papers.

\textbf{Penetration Depth (PD)} (unit: $mm$)  measures the maximum intersection distance between the object and the hand, calculated in MuJoCo. 

\textbf{Self-Penetration Depth (SPD)} (unit: $mm$) is the maximum self-intersection distance among the hand's collision meshes, ignoring the collisions between neighboring links.

\textbf{Contact Distance Consistency (CDC)} (unit: $mm$) measures the delta between the maximum and minimum signed distances across all fingers. This metric quantifies the variation in contact distance across different fingers and is invariant to penetration.

\textbf{First Variance Ratio (FVR)} (unit: $\%$) is the ratio of the first eigenvalue in PCA, indicating the proportion of variance explained by the first principal component. Each grasp data point is a grasp pose $\mathbf{x}$ with the root rotation in 3D axis-angle format, normalized by setting the object to the identity pose.

\begin{figure}[t]
    \centering
    \vspace{2mm}
    \includegraphics[width=0.85\columnwidth]{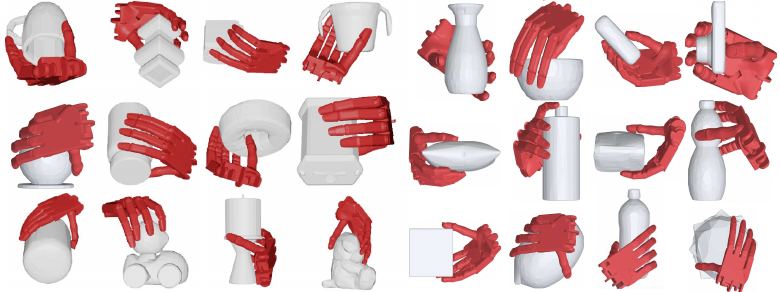}
    \caption{\textbf{More visualization of our dataset.} 
    }
    \vspace{-4mm}
    \label{fig:sim2 gallery}
\end{figure}

\subsection{Benchmarking Analytic-based Grasp Synthesis}
\label{sec:exp-analytic}
\subsubsection{Comparison with previous pipelines}
We compare with DexGraspNet (DGN)~\cite{wang2023dexgraspnet}, SpringGrasp~\cite{chen2024springgrasp}, and FRoGGeR~\cite{li2023frogger}, using the 16-DOF 4-fingered Allegro Hand, the only hand type supported by the open-source codes of all baselines. For each scaled object in a floating state without a table, $10$ grasps are synthesized, resulting in $95,880$ grasps per method. 
For baselines without pre-grasp and squeezed poses, we generate these poses by scaling the grasp poses by factors of $0.9$ and $1.1$, respectively. 
All experiments in this section follow this setting.

As shown in Table~\ref{tab:optimization} and Fig.~\ref{fig:teaser}, our pipeline greatly outperforms previous methods across nearly all metrics, especially in simulation success rate and speed. Moreover, our grasps show lower penetration depth and contact distance consistency, indicating a better contact convergence. Our diversity is also comparable to previous works. Visual comparisons are provided in Fig.~\ref{fig:sim gallery}.

\begin{table}[t]
    \vspace{2mm}
    \centering
    \begin{tabular}{|c||c|c|c|c|c|c|}
    \hline
    Name & SSR~$\uparrow$  & S~$\uparrow$  & PD~$\downarrow$  & CDC~$\downarrow$ & SPD~$\downarrow$ & FVR~$\downarrow$ \\
    \hline
    DexGraspNet & 37.0 & 0.40 & 4.49 & 5.53 & 1.30 & 32.0 \\
    SpringGrasp & 12.1 & 0.67 & 11.7 & 20.1 & 0.80 & 44.5 \\
    FRoGGeR & 17.2 & 1.20 & 2.14 & 3.96 & \textbf{0.01} & \textbf{28.9} \\
    \hline
    Ours & \textbf{79.9} & \textbf{61.5} & \textbf{0.51} & \textbf{2.98} & \textbf{0.01} & 33.2 \\
    \hline
    \end{tabular}
    \caption{\textbf{Comparison with analytic-based baselines.}}
    \vspace{-2mm}
    \label{tab:optimization}
\end{table} 

\begin{figure}[t]
    \centering
    \hspace{1mm}
    \begin{subfigure}[b]{0.47\columnwidth}
        \centering
        \includegraphics[width=\columnwidth]{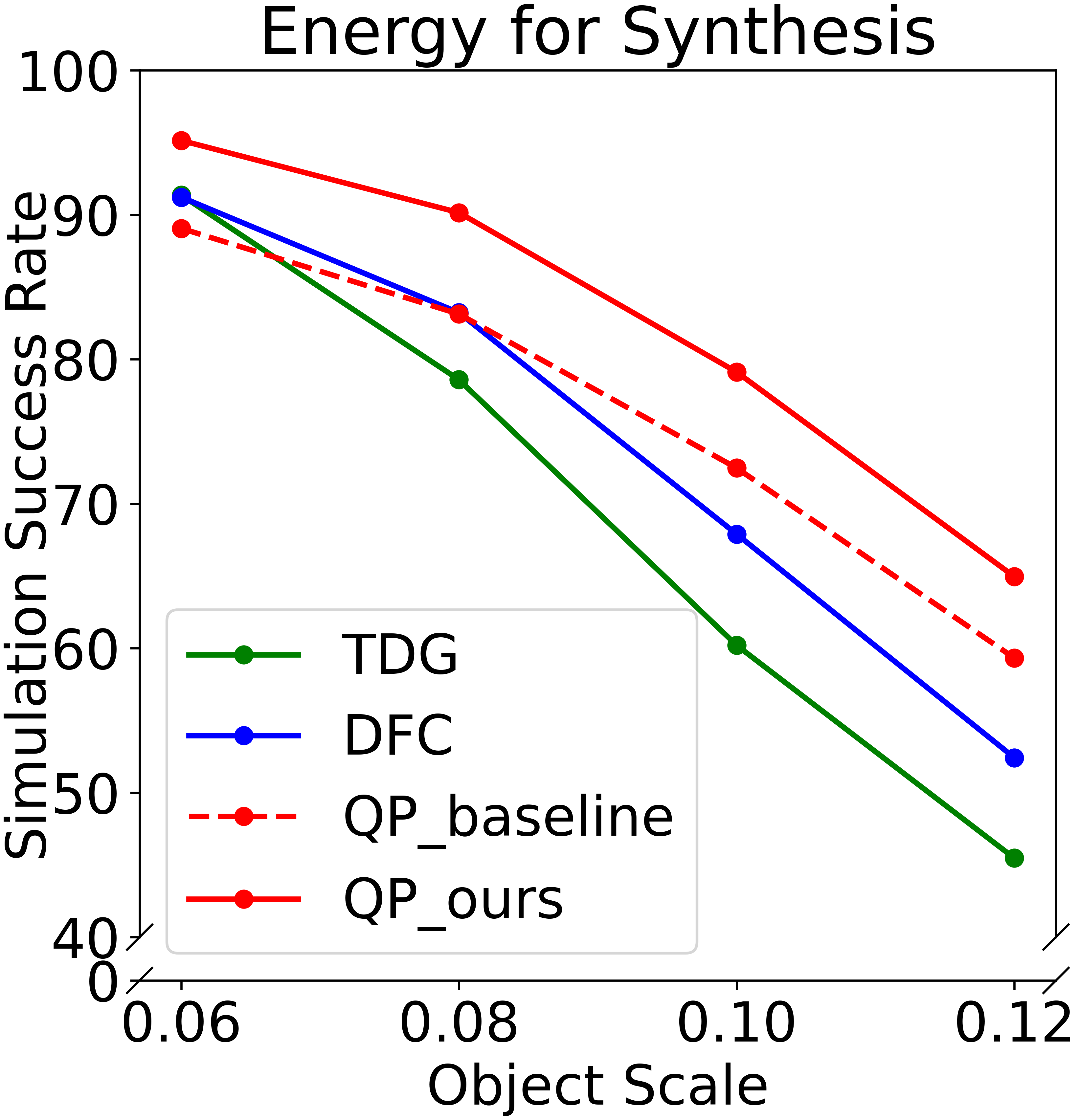}
    \end{subfigure}
    \hfill
    \begin{subfigure}[b]{0.48\columnwidth}
        \centering
        \includegraphics[width=\columnwidth]{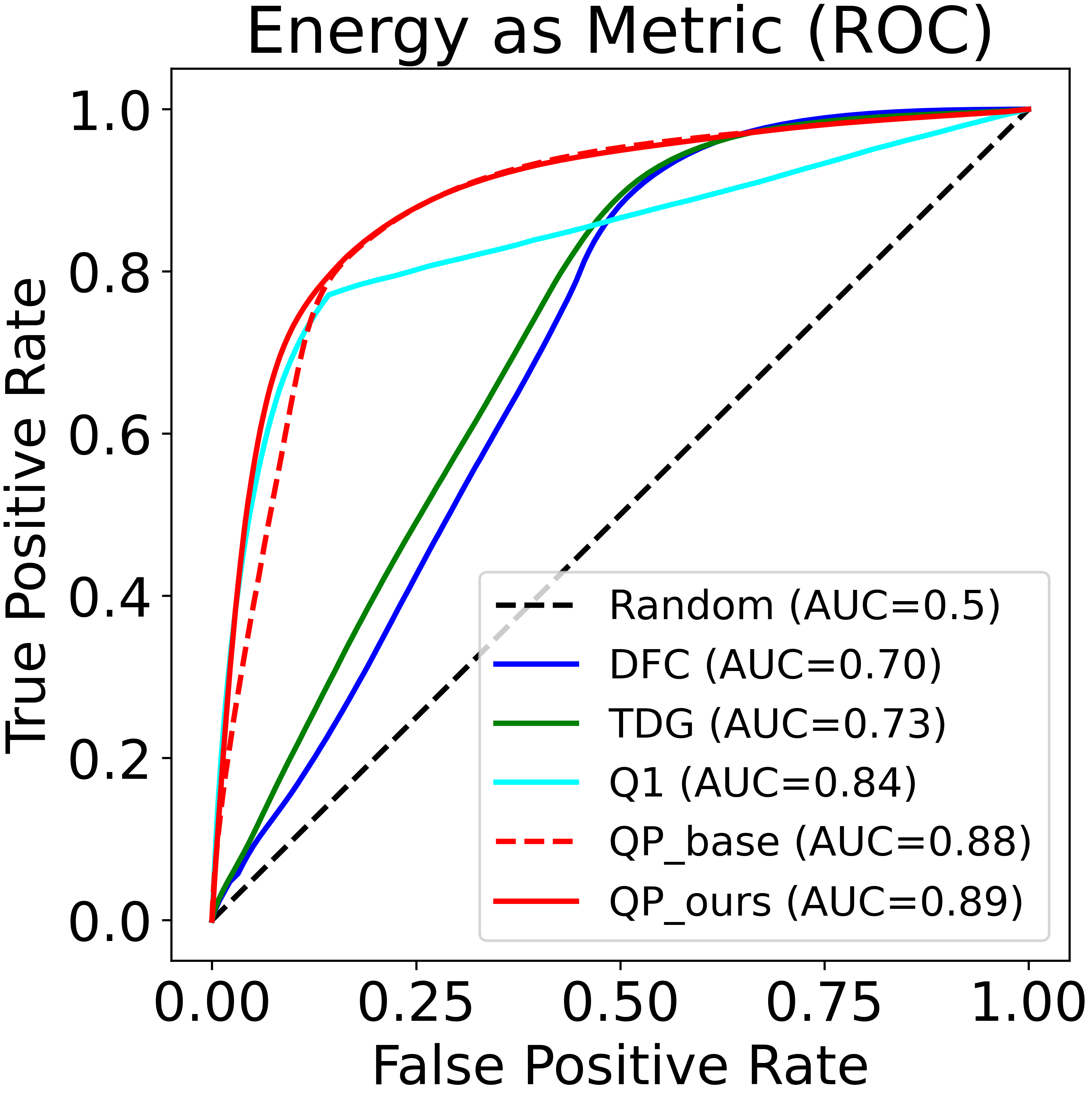}
    \end{subfigure}
    \caption{\textbf{Comparison of different grasp energy.} 
    }
    \vspace{-3mm}
    \label{fig:grasp energy}
\end{figure}

\subsubsection{Comparison with previous grasp energies}
We compare various grasp energies as objective functions for grasp synthesis and metrics for grasp evaluation. These include Q1~\cite{ferrari1992planning}, DFC~\cite{liu2021synthesizing}, TDG~\cite{chen2023task}, and QP\_baseline~\cite{wu2022learning}. QP\_baseline is similar to a special case of our method with $\beta=0$ in Eq.~\ref{eq: qp}. Here the Shadow Hand is used.

For synthesis, we re-implemented all energies within our pipeline to ensure a fair comparison. We only report the SSR metric, as other metrics are more pipeline-specific and less sensitive to the choice of grasp energy. The Q1 energy is excluded from synthesis due to its poor differentiability~\cite{liu2021synthesizing}, making it impractical for optimization-based methods.

For evaluation, we assess all grasps synthesized by the above four energies. A good grasp quality energy should ensure that grasps with lower energy are more likely to succeed in simulation. This is measured by the ROC curve, which plots the true positive rate versus the false positive rate at varying energy thresholds. The grasp energy with a larger area under the ROC curve (AUC) is better.

The results, shown in Fig.~\ref{fig:grasp energy}, demonstrate that our energy greatly outperforms DFC and TDG for both synthesis and evaluation, as our energy does not assume equal contact forces. Compared to QP\_baseline, our energy improves SSR by $10\%$ during synthesis, while the evaluation performance is comparable. Interestingly, as the object scale increases, the success rate decreases. This occurs because smaller objects are easier for the hand to form a wrapping grasp and achieve force closure, whereas larger objects make optimization harder. Notably, this decline is less pronounced with our energy, highlighting its advantages.

\begin{table}[t]
    \vspace{2mm}
    \centering
    \begin{tabular}{|c|c|c||c|c|c|c|}
    \hline
    Hand & Coarse2fine &  Pre-grasp & SSR & S & PD & CDC  \\
    \hline 
    Shadow & \checkmark & 
    \ding{55} & 57.8 & 52.7 & 0.69 & 3.07 \\
    Shadow & \ding{55} & \checkmark  & 68.4 & \textbf{55.3} & 0.81 & 3.62 \\
     Shadow & \checkmark & \checkmark  & \textbf{75.8} & 49.8 & \textbf{0.66} & \textbf{2.83}\\
    \hline
    Allegro & \checkmark & \ding{55} & 68.2 & 65.2 & 0.57 & 3.16\\
    Allegro & \ding{55} & \checkmark & 78.1 & \textbf{67.4} & 0.63 & 3.42 \\
    Allegro & \checkmark & \checkmark & \textbf{79.9} & 61.5 & \textbf{0.51} & \textbf{2.98}\\
    \hline
    \end{tabular}
     \caption{\textbf{Ablation study of our grasp synthesis pipeline.}}
    \vspace{-2mm}
    \label{tab:ablation}
\end{table}

\subsubsection{Ablation Study}
As shown in Table~\ref{tab:ablation}, the pre-grasp and coarse-to-fine strategies improve the simulation success rate, penetration depth, and contact distance consistency, at the cost of speed. Moreover, the Allegro Hand is faster than Shadow Hand due to fewer fingers.  Notably, all hyperparameters are kept the same for these two hands. We have also synthesized the grasps for Leap hand and released them on our project page, whose quantitative numbers are similar and thus not included here.

\begin{table}[t]
    \centering
    \begin{tabular}{|c|c||c|c|c|c|c|}
    \hline
    Dataset & Method & SSR~$\uparrow$ & PD~$\downarrow$ & CDC~$\downarrow$& SPD~$\downarrow$ & FVR~$\downarrow$ \\
    \hline
    Ours-F & GTTA & 14.8 & 27.3 & 29.6 & 0.15 & 39.2 \\
    Ours-F & ISAG & 20.7 & 17.3 & 22.4 & 0.08 & 38.7 \\
    Ours-F & DP3 & 62.4 & 8.91 & 13.9 & 0.10 & \textbf{32.6} \\
    Ours-F & UDG & \textbf{80.1} & \textbf{3.47} & \textbf{8.85} & \textbf{0.04} & 33.4 \\
    DGN & UDG & 42.6 & 8.69 & 12.5 & 0.13 & 33.2 \\
    \hline
    UDG & UDG & 36.8 & 12.0 & 16.5 & 0.16 & \textbf{32.7} \\
    Ours-T & UDG & \textbf{78.8} & \textbf{5.79} & \textbf{9.79} & \textbf{0.05} & 33.9 \\
    \hline
    \end{tabular}
    \caption{\textbf{Comparison of learning-based methods.}}
    \vspace{-3mm}
    \label{tab:learning}
\end{table}

\begin{figure*}[t]
    \vspace{2mm}
    \centering
    \includegraphics[width=1.98\columnwidth]{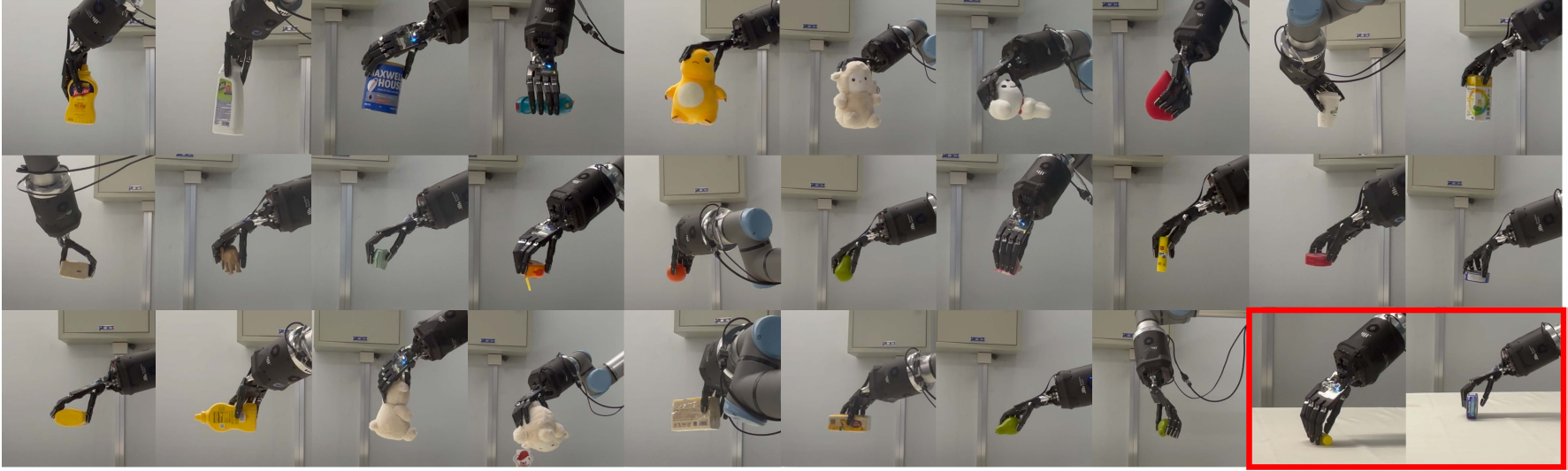}
    \caption{\textbf{Real-world grasp gallery.} All grasps are predicted by our trained network. The first two raws show one successful grasp for each object, while the last raw shows more for some objects. Two typical failure cases are shown in the red box.}
    \label{fig:gallery}
    \vspace{-3mm}
\end{figure*}

\subsection{Benchmarking Learning-based Grasp Synthesis}

We benchmark $4$ supervised learning architectures: ISAGrasp (ISAG)~\cite{chen2022learning}, GraspTTA (GTTA)~\cite{jiang2021graspTTA}, 3D Diffusion policy (DP3)~\cite{ze20243d}, and UnidexGrasp (UDG)~\cite{xu2023unidexgrasp}. They are representative due to their diverse architectures, ranging from naive regression and CVAE to diffusion models and normalizing flows. 

To ensure a fair comparison, we standardize the backbone across these methods by a 3D SparseConv network~\cite{graham2017submanifold} to extract global features from the input single-view partial point cloud. Another change is that the network learns to predict both pre-grasp and grasp pose. We also simplify the complex pipeline of UniDexGrasp and only use the GraspGlow module combined with Mobius Flow~\cite{liu2023delving} for orientation generation. For diffusion models and normalizing flows, we select the top 10 grasps from 100 samples for testing, as these models allow for probability estimation of each sample, which indicates the quality of the prediction. 

The objects are randomly split into a training set and a testing set in a 4:1 ratio. $4$ scales are applied to each object, and $5$ partial point clouds are rendered for each scaled object from random camera viewpoints. Each model is trained with $50,000$ iterations with a batch size of $256$. 

\subsubsection{Comparison with previous datasets}
To demonstrate the superiority of our dataset over previous datasets for downstream learning-based grasp synthesis, we train a similar network on different datasets for comparison. We use the official datasets from DexGraspNet and UniDexGrasp as baselines and exclude grasps that fail in MuJoCo, obtaining 356k and 238k successful grasps for training, respectively. 

The results, shown in Table~\ref{tab:learning}, indicate that models trained on our dataset consistently outperform those trained on previous datasets, demonstrating the higher quality of our dataset. Additionally, the diffusion model and normalizing flow methods perform significantly better than naive regression and CVAE, likely due to their superior expressive capabilities and the use of the top-10 selection strategy.

\subsubsection{Comparison of different dataset sizes} 
\begin{figure}[t]
    \centering
    \includegraphics[width=0.8\columnwidth]{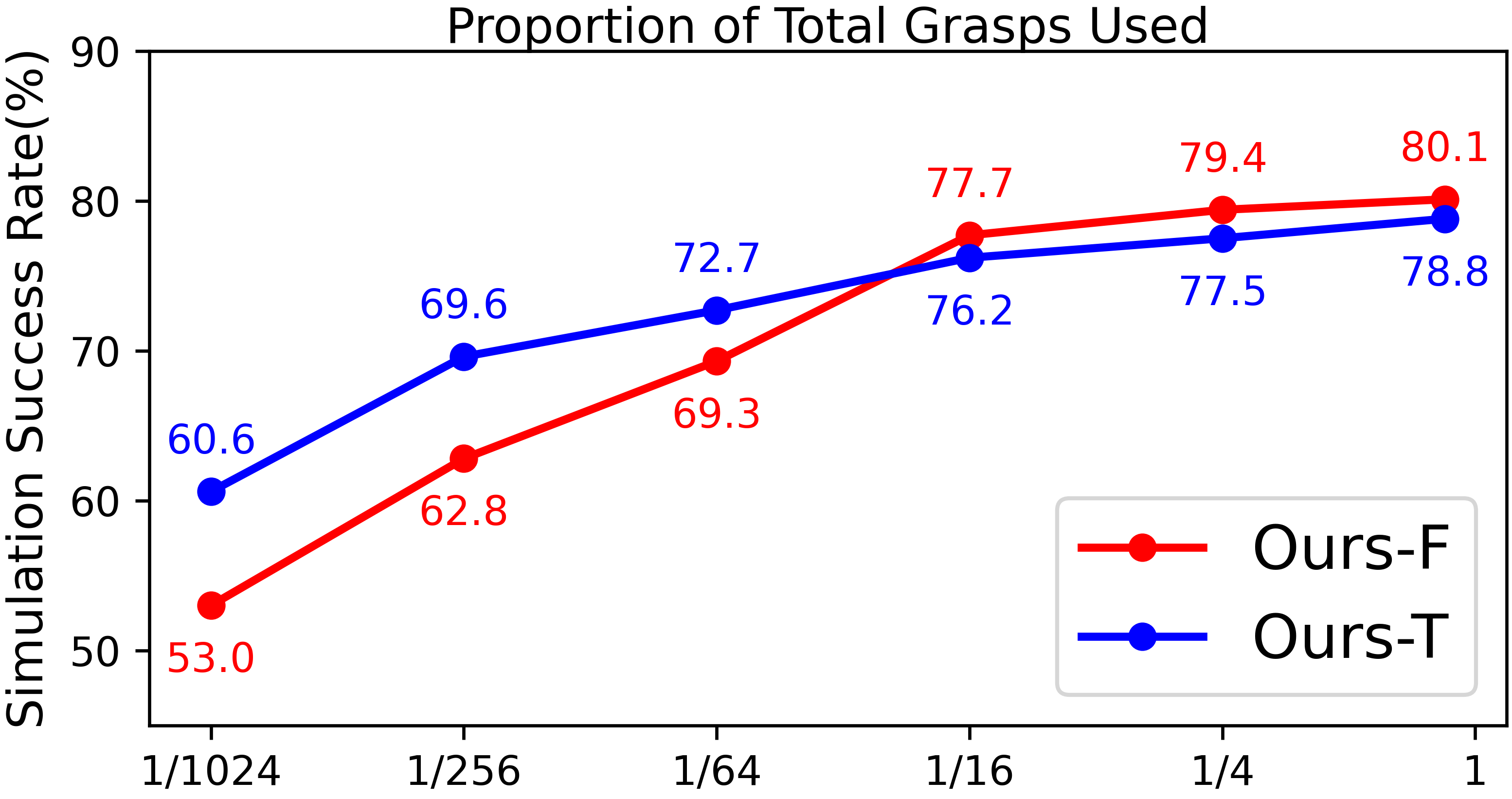}
    \caption{\textbf{Scaling the number of grasps.}}
    \vspace{-4mm}
    \label{fig:scaling}
\end{figure}

We explore the effect of dataset size on model performance, as shown in Fig.~\ref{fig:scaling}. The performance curve for datasets with objects on a table is flatter compared to that for floating objects, likely due to the more constrained distribution of the hand root pose—restricted to positions over the table. The results also show that increasing the number of grasps in the training dataset consistently improves performance, though the gains slow down once the dataset scale reaches the million level. Future work could explore enhancing model capacity to better utilize larger datasets.

\subsection{Real-World Experiments}

\begin{figure}[t]
    \centering
    \includegraphics[width=0.9\columnwidth]{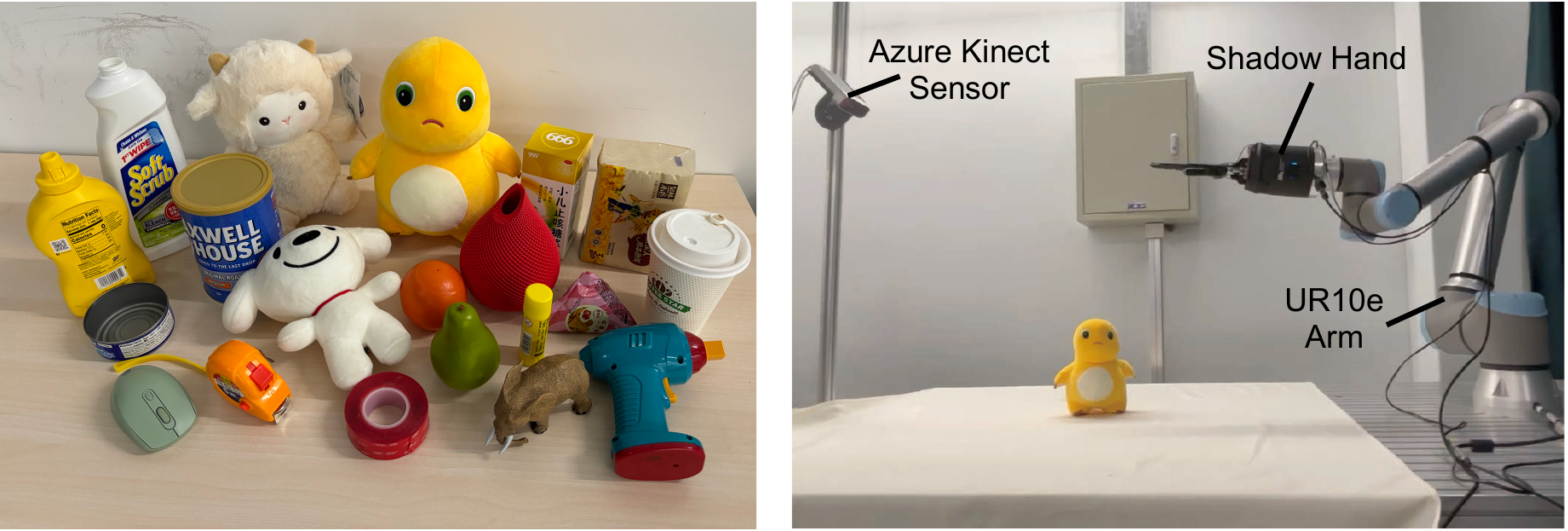}
    \caption{\textbf{Real-world experiment setup.} \textit{Left}: 20 test objects.}
    \label{fig:real_world}
    \vspace{-4mm}
\end{figure}

Finally, we validate the best-performing trained model in the real world. The hardware setup includes a 22-DoF Shadow Hand mounted on a 6-DoF UR10e robotic arm, and an Azure Kinect sensor to capture the RGB and depth images, as shown in Fig.~\ref{fig:real_world}. The experiments are conducted on 20 objects, with 5 grasp attempts per object.

For each trial, we first segment the object in the RGB image using Segment Anything~\cite{ravi2024sam} and feed the segmented depth point cloud into the trained model. The network then outputs 100 pairs of pre-grasp and grasp poses. Each pre-grasp pose is sequentially used as the target for collision-free motion planning with cuRobo, ordered by their probabilities. The first successfully planned pose is executed. Due to the noise in camera calibration, depth sensing, and network output, we find that a $1cm$ margin is insufficient in the real world, so we instead use $\mathbf{x}_p'=2\mathbf{x}_p-\mathbf{x}$ as the new pre-grasp for motion planning. Finally, the hand moves to the squeezed pose $\mathbf{x}_s$ to apply force to the object before lifting it.

Our trained model achieves an overall success rate of $81\%$. As shown in Fig.~\ref{fig:gallery}, it successfully grasps both large objects, such as bottles and toys in the first row, as well as thin and flat objects, like the last three in the second row.

We also observe two typical failure cases, which tend to occur more frequently with thin and flat objects. In one scenario, the predicted grasp misses the object by a small margin; in the other, the predicted grasp is too wide and requires additional squeezing. Improving these cases may require incorporating more in-domain data during training.

\section{Limitations and Future Work}
First, our pipeline requires designating contact spheres on the hand, which we have placed on each fingertip. So the generated grasps rely solely on fingertips and do not utilize the palm. While fingertip-only grasps may facilitate future studies on tactile feedback, they lack the robustness that palm contact provides. Second, our work does not study grasping in a cluster scene. Third, our generated grasps are not functional. Finally, the collision-free trajectories in our dataset are not currently utilized, which may enable the training of a closed-loop visual policy in the future. 

\section{Conclusions}
In this work, we present a scalable and efficient pipeline for robotic dexterous grasp synthesis, designed to facilitate the construction of large-scale, high-quality datasets and enhance data-driven grasp synthesis methods. We also establish a benchmark with MuJoCo to compare with previous approaches, demonstrating the superiority of both our pipeline and dataset. Real-world experiments further validate our effectiveness and its potential for future applications.

\section{Acknowledgment}
This work was supported by Beijing Natural Science Foundation (Grant No.QY24042).

\addtolength{\textheight}{-12cm}   




\bibliographystyle{IEEEtran}
\bibliography{IEEEabrv}

\begin{thebibliography}{10}
\providecommand{\url}[1]{#1}
\csname url@rmstyle\endcsname
\providecommand{\newblock}{\relax}
\providecommand{\bibinfo}[2]{#2}
\providecommand\BIBentrySTDinterwordspacing{\spaceskip=0pt\relax}
\providecommand\BIBentryALTinterwordstretchfactor{4}
\providecommand\BIBentryALTinterwordspacing{\spaceskip=\fontdimen2\font plus
\BIBentryALTinterwordstretchfactor\fontdimen3\font minus \fontdimen4\font\relax}
\providecommand\BIBforeignlanguage[2]{{%
\expandafter\ifx\csname l@#1\endcsname\relax
\typeout{** WARNING: IEEEtran.bst: No hyphenation pattern has been}%
\typeout{** loaded for the language `#1'. Using the pattern for}%
\typeout{** the default language instead.}%
\else
\language=\csname l@#1\endcsname
\fi
#2}}

\bibitem{fang2020graspnet}
H.-S. Fang, C.~Wang, M.~Gou, and C.~Lu, ``Graspnet-1billion: A large-scale benchmark for general object grasping,'' in \emph{Proceedings of the IEEE/CVF conference on computer vision and pattern recognition}, 2020, pp. 11\,444--11\,453.

\bibitem{fang2023anygrasp}
H.-S. Fang, C.~Wang, H.~Fang, M.~Gou, J.~Liu, H.~Yan, W.~Liu, Y.~Xie, and C.~Lu, ``Anygrasp: Robust and efficient grasp perception in spatial and temporal domains,'' \emph{IEEE Transactions on Robotics}, 2023.

\bibitem{liu2021synthesizing}
T.~Liu, Z.~Liu, Z.~Jiao, Y.~Zhu, and S.-C. Zhu, ``Synthesizing diverse and physically stable grasps with arbitrary hand structures using differentiable force closure estimator,'' \emph{IEEE Robotics and Automation Letters}, vol.~7, no.~1, pp. 470--477, 2021.

\bibitem{wang2023dexgraspnet}
R.~Wang, J.~Zhang, J.~Chen, Y.~Xu, P.~Li, T.~Liu, and H.~Wang, ``Dexgraspnet: A large-scale robotic dexterous grasp dataset for general objects based on simulation,'' in \emph{2023 IEEE International Conference on Robotics and Automation (ICRA)}.\hskip 1em plus 0.5em minus 0.4em\relax IEEE, 2023, pp. 11\,359--11\,366.

\bibitem{li2023frogger}
A.~H. Li, P.~Culbertson, J.~W. Burdick, and A.~D. Ames, ``Frogger: Fast robust grasp generation via the min-weight metric,'' \emph{arXiv preprint arXiv:2302.13687}, 2023.

\bibitem{chen2024springgrasp}
S.~Chen, J.~Bohg, and C.~K. Liu, ``Springgrasp: An optimization pipeline for robust and compliant dexterous pre-grasp synthesis,'' \emph{arXiv preprint arXiv:2404.13532}, 2024.

\bibitem{curobo_report23}
B.~Sundaralingam, S.~K.~S. Hari, A.~Fishman, C.~Garrett, K.~V. Wyk, V.~Blukis, A.~Millane, H.~Oleynikova, A.~Handa, F.~Ramos, N.~Ratliff, and D.~Fox, ``curobo: Parallelized collision-free minimum-jerk robot motion generation,'' 2023.

\bibitem{bishop2024relu}
A.~L. Bishop, J.~Z. Zhang, S.~Gurumurthy, K.~Tracy, and Z.~Manchester, ``Relu-qp: A gpu-accelerated quadratic programming solver for model-predictive control,'' in \emph{2024 IEEE International Conference on Robotics and Automation (ICRA)}.\hskip 1em plus 0.5em minus 0.4em\relax IEEE, 2024, pp. 13\,285--13\,292.

\bibitem{bambade2023proxqp}
A.~Bambade, F.~Schramm, S.~El~Kazdadi, S.~Caron, A.~Taylor, and J.~Carpentier, ``Proxqp: an efficient and versatile quadratic programming solver for real-time robotics applications and beyond,'' 2023.

\bibitem{stellato2020osqp}
B.~Stellato, G.~Banjac, P.~Goulart, A.~Bemporad, and S.~Boyd, ``Osqp: An operator splitting solver for quadratic programs,'' \emph{Mathematical Programming Computation}, vol.~12, no.~4, pp. 637--672, 2020.

\bibitem{liu2020deep}
M.~Liu, Z.~Pan, K.~Xu, K.~Ganguly, and D.~Manocha, ``Deep differentiable grasp planner for high-dof grippers,'' \emph{arXiv preprint arXiv:2002.01530}, 2020.

\bibitem{li2023gendexgrasp}
P.~Li, T.~Liu, Y.~Li, Y.~Geng, Y.~Zhu, Y.~Yang, and S.~Huang, ``Gendexgrasp: Generalizable dexterous grasping,'' in \emph{2023 IEEE International Conference on Robotics and Automation (ICRA)}.\hskip 1em plus 0.5em minus 0.4em\relax IEEE, 2023, pp. 8068--8074.

\bibitem{liu2024realdex}
Y.~Liu, Y.~Yang, Y.~Wang, X.~Wu, J.~Wang, Y.~Yao, S.~Schwertfeger, S.~Yang, W.~Wang, J.~Yu, \emph{et~al.}, ``Realdex: Towards human-like grasping for robotic dexterous hand,'' \emph{arXiv preprint arXiv:2402.13853}, 2024.

\bibitem{todorov2012mujoco}
E.~Todorov, T.~Erez, and Y.~Tassa, ``Mujoco: A physics engine for model-based control,'' in \emph{2012 IEEE/RSJ international conference on intelligent robots and systems}.\hskip 1em plus 0.5em minus 0.4em\relax IEEE, 2012, pp. 5026--5033.

\bibitem{ferrari1992planning}
C.~Ferrari, J.~F. Canny, \emph{et~al.}, ``Planning optimal grasps.'' in \emph{ICRA}, vol.~3, no.~4, 1992, p.~6.

\bibitem{miller2004graspit}
A.~T. Miller and P.~K. Allen, ``Graspit! a versatile simulator for robotic grasping,'' \emph{IEEE Robotics \& Automation Magazine}, vol.~11, no.~4, pp. 110--122, 2004.

\bibitem{turpin2022grasp}
D.~Turpin, L.~Wang, E.~Heiden, Y.-C. Chen, M.~Macklin, S.~Tsogkas, S.~Dickinson, and A.~Garg, ``Grasp’d: Differentiable contact-rich grasp synthesis for multi-fingered hands,'' in \emph{European Conference on Computer Vision}.\hskip 1em plus 0.5em minus 0.4em\relax Springer, 2022, pp. 201--221.

\bibitem{turpin2023fast}
D.~Turpin, T.~Zhong, S.~Zhang, G.~Zhu, J.~Liu, R.~Singh, E.~Heiden, M.~Macklin, S.~Tsogkas, S.~Dickinson, \emph{et~al.}, ``Fast-grasp'd: Dexterous multi-finger grasp generation through differentiable simulation,'' \emph{arXiv preprint arXiv:2306.08132}, 2023.

\bibitem{chen2023task}
J.~Chen, Y.~Chen, J.~Zhang, and H.~Wang, ``Task-oriented dexterous grasp synthesis via differentiable grasp wrench boundary estimator,'' \emph{arXiv preprint arXiv:2309.13586}, 2023.

\bibitem{wu2022learning}
A.~Wu, M.~Guo, and C.~K. Liu, ``Learning diverse and physically feasible dexterous grasps with generative model and bilevel optimization,'' \emph{arXiv preprint arXiv:2207.00195}, 2022.

\bibitem{jiang2021graspTTA}
H.~Jiang, S.~Liu, J.~Wang, and X.~Wang, ``Hand-object contact consistency reasoning for human grasps generation,'' in \emph{Proceedings of the International Conference on Computer Vision}, 2021.

\bibitem{chen2022learning}
Z.~Q. Chen, K.~Van~Wyk, Y.-W. Chao, W.~Yang, A.~Mousavian, A.~Gupta, and D.~Fox, ``Learning robust real-world dexterous grasping policies via implicit shape augmentation,'' \emph{arXiv preprint arXiv:2210.13638}, 2022.

\bibitem{xu2023unidexgrasp}
Y.~Xu, W.~Wan, J.~Zhang, H.~Liu, Z.~Shan, H.~Shen, R.~Wang, H.~Geng, Y.~Weng, J.~Chen, \emph{et~al.}, ``Unidexgrasp: Universal robotic dexterous grasping via learning diverse proposal generation and goal-conditioned policy,'' in \emph{Proceedings of the IEEE/CVF Conference on Computer Vision and Pattern Recognition}, 2023, pp. 4737--4746.

\bibitem{kingma2013auto}
D.~P. Kingma, ``Auto-encoding variational bayes,'' \emph{arXiv preprint arXiv:1312.6114}, 2013.

\bibitem{ho2020denoising}
J.~Ho, A.~Jain, and P.~Abbeel, ``Denoising diffusion probabilistic models,'' \emph{Advances in neural information processing systems}, vol.~33, pp. 6840--6851, 2020.

\bibitem{rezende2015variational}
D.~Rezende and S.~Mohamed, ``Variational inference with normalizing flows,'' in \emph{International conference on machine learning}.\hskip 1em plus 0.5em minus 0.4em\relax PMLR, 2015, pp. 1530--1538.

\bibitem{wan2023unidexgrasp++}
W.~Wan, H.~Geng, Y.~Liu, Z.~Shan, Y.~Yang, L.~Yi, and H.~Wang, ``Unidexgrasp++: Improving dexterous grasping policy learning via geometry-aware curriculum and iterative generalist-specialist learning,'' \emph{arXiv preprint arXiv:2304.00464}, 2023.

\bibitem{schulman2014motion}
J.~Schulman, Y.~Duan, J.~Ho, A.~Lee, I.~Awwal, H.~Bradlow, J.~Pan, S.~Patil, K.~Goldberg, and P.~Abbeel, ``Motion planning with sequential convex optimization and convex collision checking,'' \emph{The International Journal of Robotics Research}, vol.~33, no.~9, pp. 1251--1270, 2014.

\bibitem{gilbert1988fast}
E.~G. Gilbert, D.~W. Johnson, and S.~S. Keerthi, ``A fast procedure for computing the distance between complex objects in three-dimensional space,'' \emph{IEEE Journal on Robotics and Automation}, vol.~4, no.~2, pp. 193--203, 1988.

\bibitem{Pan_Coal_-_An_2025}
\BIBentryALTinterwordspacing
J.~Pan, S.~Chitta, J.~Pan, D.~Manocha, J.~Mirabel, J.~Carpentier, and L.~Montaut, ``{Coal - An extension of the Flexible Collision Library},'' Feb. 2025. [Online]. Available: \url{https://github.com/coal-library/coal}
\BIBentrySTDinterwordspacing

\bibitem{menagerie2022github}
\BIBentryALTinterwordspacing
K.~Zakka, Y.~Tassa, and {MuJoCo Menagerie Contributors}, ``{MuJoCo Menagerie: A collection of high-quality simulation models for MuJoCo},'' 2022. [Online]. Available: \url{http://github.com/google-deepmind/mujoco\_menagerie}
\BIBentrySTDinterwordspacing

\bibitem{xydas1999modeling}
N.~Xydas and I.~Kao, ``Modeling of contact mechanics and friction limit surfaces for soft fingers in robotics, with experimental results,'' \emph{The International Journal of Robotics Research}, vol.~18, no.~9, pp. 941--950, 1999.

\bibitem{ze20243d}
Y.~Ze, G.~Zhang, K.~Zhang, C.~Hu, M.~Wang, and H.~Xu, ``3d diffusion policy,'' \emph{arXiv preprint arXiv:2403.03954}, 2024.

\bibitem{graham2017submanifold}
B.~Graham and L.~Van~der Maaten, ``Submanifold sparse convolutional networks,'' \emph{arXiv preprint arXiv:1706.01307}, 2017.

\bibitem{liu2023delving}
Y.~Liu, H.~Liu, Y.~Yin, Y.~Wang, B.~Chen, and H.~Wang, ``Delving into discrete normalizing flows on so (3) manifold for probabilistic rotation modeling,'' in \emph{Proceedings of the IEEE/CVF Conference on Computer Vision and Pattern Recognition}, 2023, pp. 21\,264--21\,273.

\bibitem{ravi2024sam}
N.~Ravi, V.~Gabeur, Y.-T. Hu, R.~Hu, C.~Ryali, T.~Ma, H.~Khedr, R.~R{\"a}dle, C.~Rolland, L.~Gustafson, \emph{et~al.}, ``Sam 2: Segment anything in images and videos,'' \emph{arXiv preprint arXiv:2408.00714}, 2024.

\end{thebibliography}

\end{document}